\documentclass[conference]{IEEEtran}
\IEEEoverridecommandlockouts
\usepackage{cite}
\usepackage{amsmath,amssymb,amsfonts}
\usepackage{algorithmic}
\usepackage{graphicx}
\usepackage{textcomp}
\usepackage{xcolor}
\usepackage{times}
\usepackage{epsfig}
\usepackage{amssymb}
\usepackage{booktabs}
\usepackage{tabularx}
\usepackage{color}
\usepackage{colortbl}
\usepackage{subfigure}
\usepackage{multirow}
\usepackage{comment}
\usepackage{adjustbox}

\definecolor{red}{rgb}{1,0,0}
\definecolor{blue}{rgb}{0,0,1}
\definecolor{purple}{rgb}{1,0,1}

\makeatletter
\newcommand{\linebreakand}{
  \end{@IEEEauthorhalign}
  \hfill\mbox{}\par
  \mbox{}\hfill\begin{@IEEEauthorhalign}
}
\makeatother

\def\BibTeX{{\rm B\kern-.05em{\sc i\kern-.025em b}\kern-.08em
    T\kern-.1667em\lower.7ex\hbox{E}\kern-.125emX}}

\begin{document}

\title{Context-Preserving Instance-Level Augmentation \\ and Deformable Convolution Networks \\for SAR Ship Detection}

\author{
\IEEEauthorblockN{Taeyong Song}
\IEEEauthorblockA{\textit{School of Electrical and} \\
\textit{Electronic Engineering} \\
\textit{Yonsei University}\\
Seoul, Korea \\
e-mail: sty37@yonsei.ac.kr}
\and
\IEEEauthorblockN{Sunok Kim}
\IEEEauthorblockA{\textit{Department of Software Engineering} \\
\textit{Korea Aerospace University}\\
Goyang, Korea \\
e-mail: sunok.kim@kau.ac.kr\\}
\and
\IEEEauthorblockN{SungTai Kim}
\IEEEauthorblockA{\textit{Radar Research and} \\
\textit{Development Center} \\
\textit{Hanwha Systems}\\
Yongin, Korea \\
e-mail: st2002.kim@hanwha.com}
\linebreakand 
\IEEEauthorblockN{Jaeseok Lee}
\IEEEauthorblockA{\textit{Radar Research and} \\
\textit{Development Center} \\
\textit{Hanwha Systems}\\
Yongin, Korea \\
e-mail: jaeseoklee@hanwha.com}
\and
\IEEEauthorblockN{Kwanghoon Sohn}
\IEEEauthorblockA{\textit{School of Electrical and} \\
\textit{Electronic Engineering} \\
\textit{Yonsei University}\\
Seoul, Korea \\
e-mail: khsohn@yonsei.ac.kr}
}
\vspace{-30pt}
\maketitle

\begin{abstract}
Shape deformation of targets in SAR image due to random orientation and partial information loss caused by occlusion of the radar signal, is an essential challenge in SAR ship detection.
In this paper, we propose a data augmentation method to train a deep network that is robust to partial information loss within the targets.
Taking advantage of ground-truth annotations for bounding box and instance segmentation mask, we present a simple and effective pipeline to simulate information loss on targets in instance-level, while preserving contextual information.
Furthermore, we adopt deformable convolutional network to adaptively extract shape-invariant deep features from geometrically translated targets.
By learning sampling offset to the grid of standard convolution, the network can robustly extract the features from targets with shape variations for SAR ship detection.
Experiments on the HRSID dataset including comparisons with other deep networks and augmentation methods, as well as ablation study, demonstrate the effectiveness of our proposed method.
\end{abstract}


\section{Introduction}
\label{sec:intro}
Synthetic Aperture Radar (SAR) is an airborne or satellite imaging system that uses active radar signal.
Due to its penetration capacity from spectral characteristics of radar signal, SAR image can consistently provide information under various conditions such as low illumination and adverse weather such as clouds, enabling all-day and all-weather applications.
With the advantages of capability in capturing surface information including geometry and material characteristics, SAR images are widely used for various remote sensing applications, including land cover classification \cite{liu2019polsf}, change detection \cite{rignot1993change}, building segmentation \cite{shermeyer2020spacenet}, and image recognition such as target classification \cite{keydel1996mstar} and detection \cite{li2017ship, wei2020hrsid}.

\begin{figure}[t]
	\centering
	{\includegraphics[width=1\columnwidth]{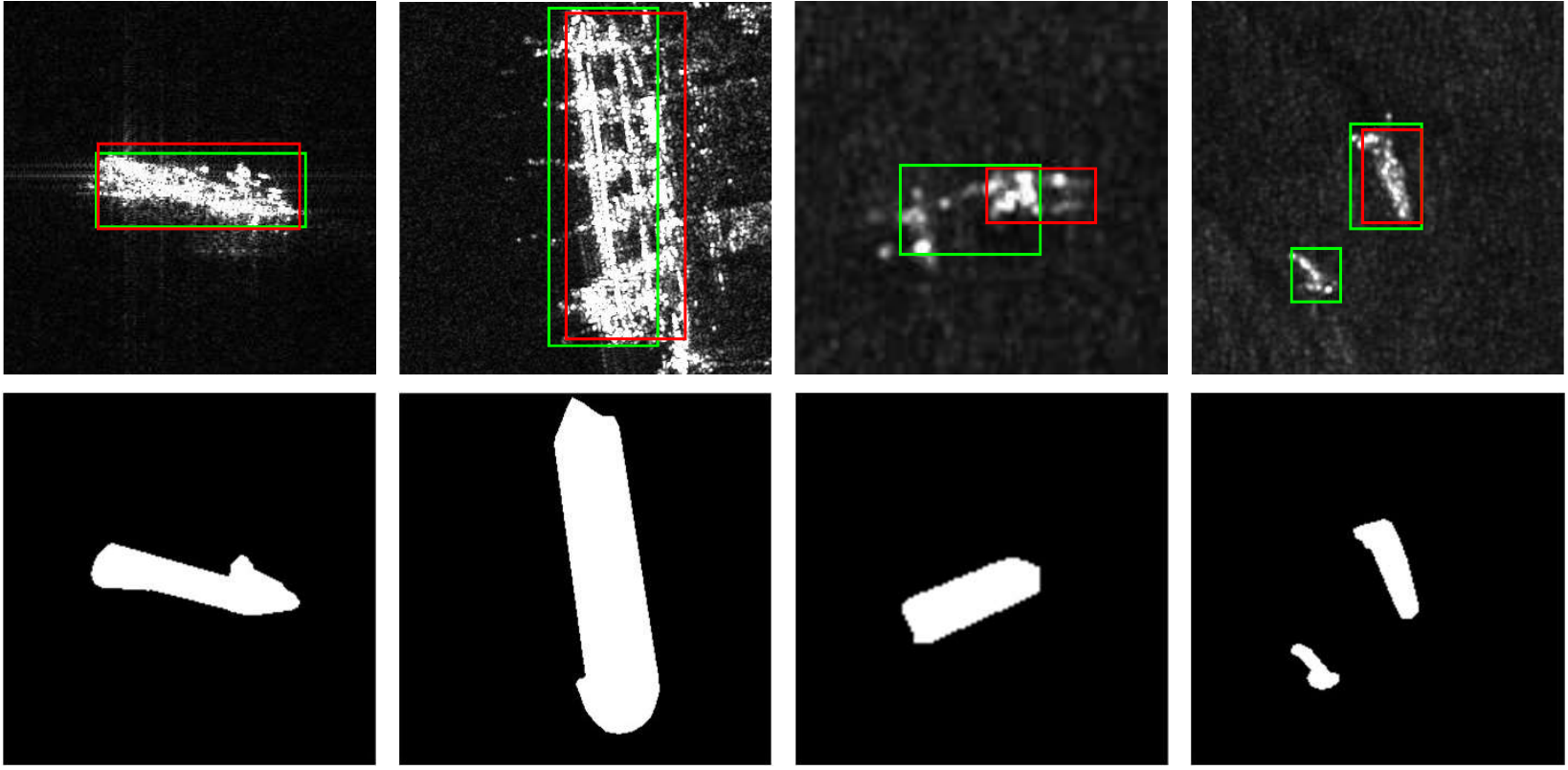}}\vspace{-5pt}
	\caption{Ship instances in the HRSID dataset \cite{wei2020hrsid}. 
	(top) Images, ground-truth bounding boxes (green) and detection results of Faster R-CNN \cite{ren2015faster} (red), 
	(bottom) instance segmentation mask.
	The network fails to detect ships when they have occluded region (3rd row) or appear in uncommon shape (4th row).}
	\vspace{-10pt}
	\label{fig:intro}
\end{figure}

\begin{figure*}[t]
	\centering
	{\includegraphics[width=1.0\textwidth]{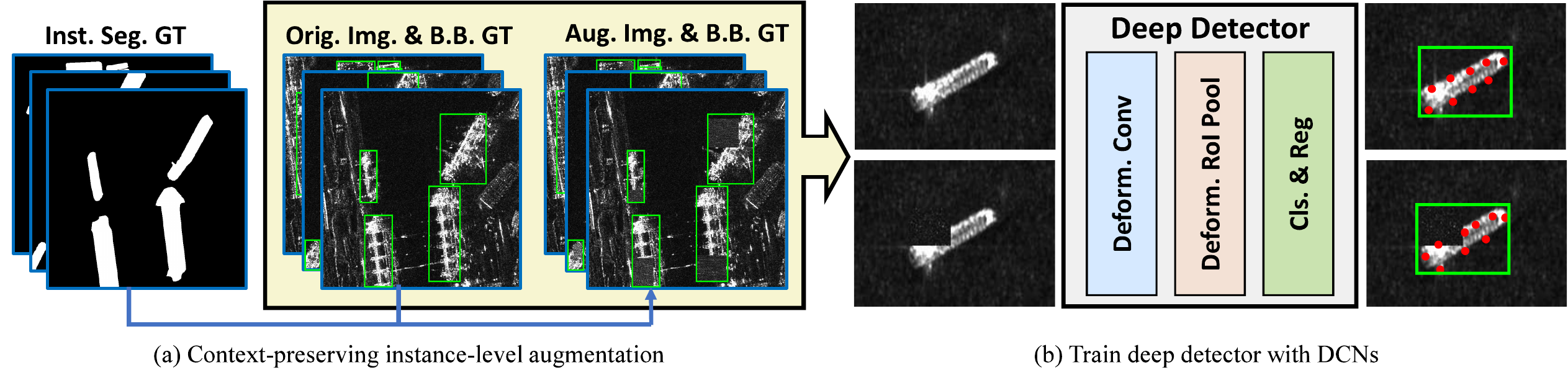}}\vspace{-8pt}
	\caption{Overall framework of our proposed framework. (a) We build the augmented dataset by exploiting the ground-truth annotations for bounding box and instance segmentation mask, then
	(b) we train a deep detector with DCN for ship detection using the augmented dataset.} 
	\vspace{-8pt}
	\label{fig:Framework}
\end{figure*}
With the establishment of large-scale datasets and improved hardware, developments of Convolutional Neural Networks (CNNs) have achieved great success in various image recognition applications in computer vision society \cite{krizhevsky2012imagenet, he2016deep, ren2015faster, farhadi2018yolov3, long2015fully}.
Success of CNNs in computer vision has encouraged researchers to solve the SAR image recognition tasks with CNN-based approaches.
As an early work, Chen et al. \cite{chen2014sar} proposed to generate a convolution kernel using sparse auto-encoder, which is later used in a single-layer CNN.
In \cite{du2016sar}, they tackled the assumption of precise location in previous SAR target classification algorithms, and proposed data augmentation strategy to train a rotation- and displacement-insensitive CNN.
A multi-task learning approach for SAR target classification is proposed by Wang et al. \cite{wang2020deep}, that simultaneously performs target classification and segmentation to achieve improved performance.
Wei et al., \cite{wei2020hrsid} constructed a high-resolution SAR dataset for deep-learning based ship detection and instance segmentation, and also presented baseline results of various deep object detectors.

Meanwhile, sufficiently large and variety of dataset is considered as an essential component for the generalization ability and robustness of CNNs.
A network trained with an insufficient dataset may suffer from severe performance degradation when fed with data whose characteristics are different from the training dataset.
To tackle this issue, many data augmentation approaches have been proposed.
Zhang et al. \cite{zhang2017mixup} proposed to train deep networks on convex combinations of pairs of examples and their labels and increased the robustness to adversarial examples.
There have been proposed a few methods \cite{devries2017improved, zhong2020random} that are closely related to ours.
They proposed to augment dataset by erasing random image regions and filling with consistent \cite{devries2017improved} or random \cite{zhong2020random} value.
Yun et al. \cite{yun2019cutmix} proposed CutMix strategy which cut and paste patches among training images, as well as mixed the ground-truth labels proportionally to areas of the patches.

On the other hands, in SAR images, targets can be presented with shape deformations due to physical characteristics of SAR images, which may lead to performance degradation in recognition tasks, as exemplified in Fig. \ref{fig:intro}.
Since SAR images are taken from bird-eye view, the targets can be oriented to an arbitrary direction.
Furthermore, radar shadow, \emph{i.e.} regions where the radar signal is occluded, can cause additional deformations and information loss.
There have been proposed several methods to alleviate the problem.
Zhu et al. \cite{zhu2020ground} proposed to restore the occluded part of SAR targets using corresponding simulated optical images by establishing relationship between their pixels.
They used the filled images and achieved better classification results with various models.
In \cite{zhu2020shadow}, they learned a variant of generative adversarial network \cite{goodfellow2014generative} to generate un-occluded SAR target and trained subsequent CNN classifier using the un-occluded target profiles and achieved improved performance.
He et al. \cite{he2019adaptive} proposed adaptive weighting strategy based on sub-image sparse model to suppress reconstruction error at the occluded area.
In \cite{he2020fusion}, they have proposed to calculate sparse model reconstruction errors using multiple randomly erased sample-dictionary pairs, then performed decision-level fusion to eliminate error caused by occlusion.

\begin{figure}[t]
\centering
{\includegraphics[width=1\columnwidth]{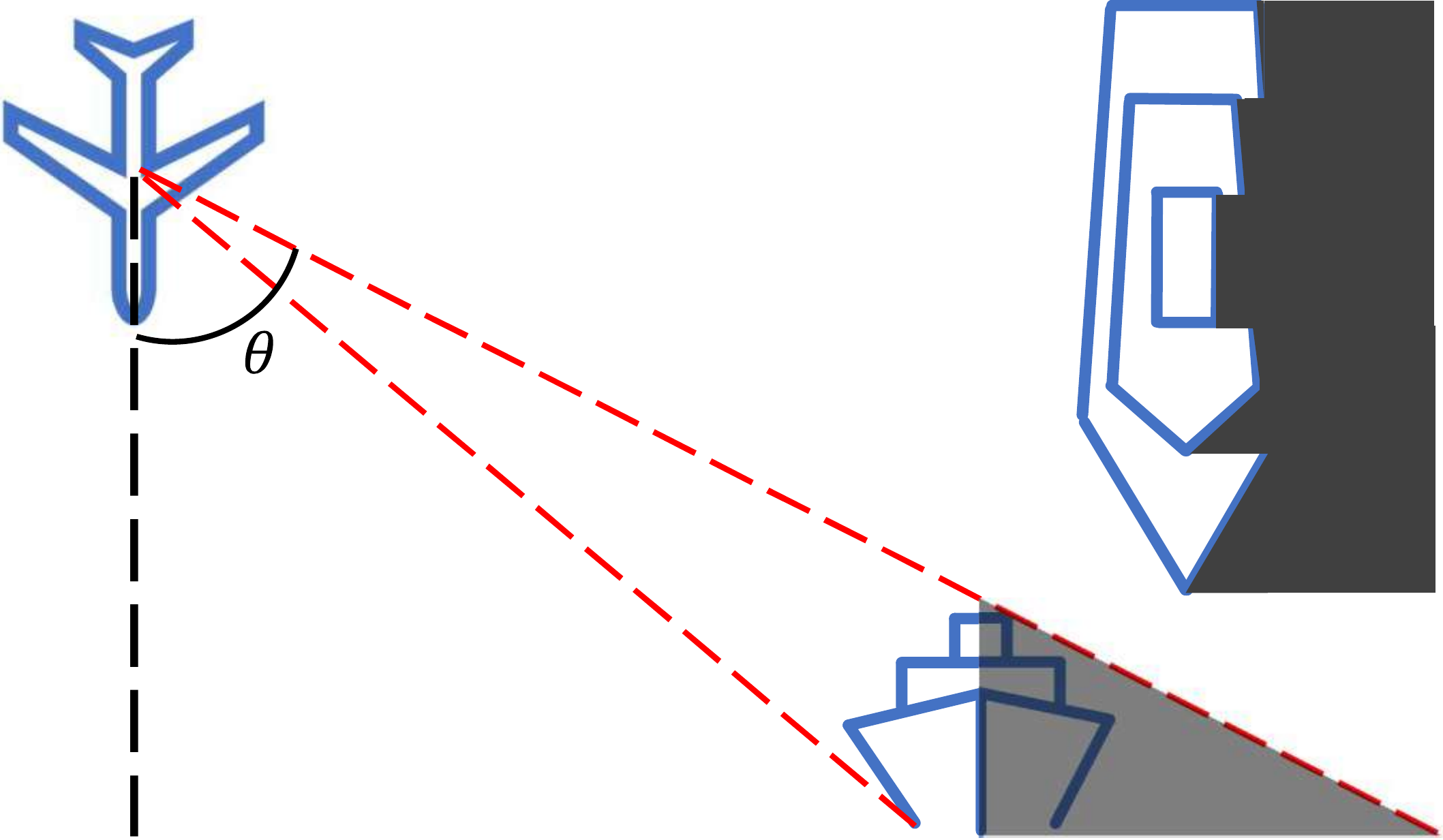}}\vspace{-5pt}
\caption{Illustration of geometries and occurrence of radar shadow in SAR image acquisition.
Due to a slanted incidence angle $\theta$, no reflection of the radar signal is received from the shaded area and limits full target information.}
\vspace{-15pt}
\label{fig:shadow}
\end{figure}
In this paper, we aim to train a deep network for SAR ship detection that is robust to target shape variations.
To this end, we propose a simple yet effective instance-level data augmentation method.
Taking advantage of the bounding box and instance segmentation mask annotations, we simulate information loss within targets without harming contextual information.
In addition, we adopt deformable convolutional network \cite{dai2017deformable} to further improve the performance.
By learning the convolution sampling offsets, the network learns to adaptively capture the information from the targets with shape variations. 
Experimental results on HRSID dataset \cite{wei2020hrsid} demonstrate the effectiveness of the proposed method.

\section{Proposed Method}
\label{sec:proposed}
    \subsection{Problem Formulation and Overview}
        Let us consider a SAR image $\mathbf{I}$ that includes $N_s$ ship instances. 
        Each instance is provided with corresponding ground-truth bounding box $\mathbf{b}_n$ by coordinates of the upper left corner, width, and height as $\mathbf{b}_n=[x,y,w,h]$, and binary instance segmentation mask $\mathbf{M}_n$ for each instance, where $n={1,2,...,N_s}$ is index of each ship instance.
        For SAR ship detection, we aim to find the location of each ship instances.
        To realize this, we design a novel deep-learning framework for SAR ship detection that is robust to various shape variations and occlusions. 
        We propose a context-preserving instance-level data augmentation method to deal with shape variations that accompanies target information loss (Section \ref{sec:proposed_aug}). 
        To deal with the various shape variation, we apply deformable convolutional network \cite{dai2017deformable} to extract shape-adaptive deep features (Section \ref{sec:proposed_network}).
        Fig. 2 shows the overall configuration of our proposed framework.

\begin{figure}[t]
\centering
{\includegraphics[width=1\columnwidth]{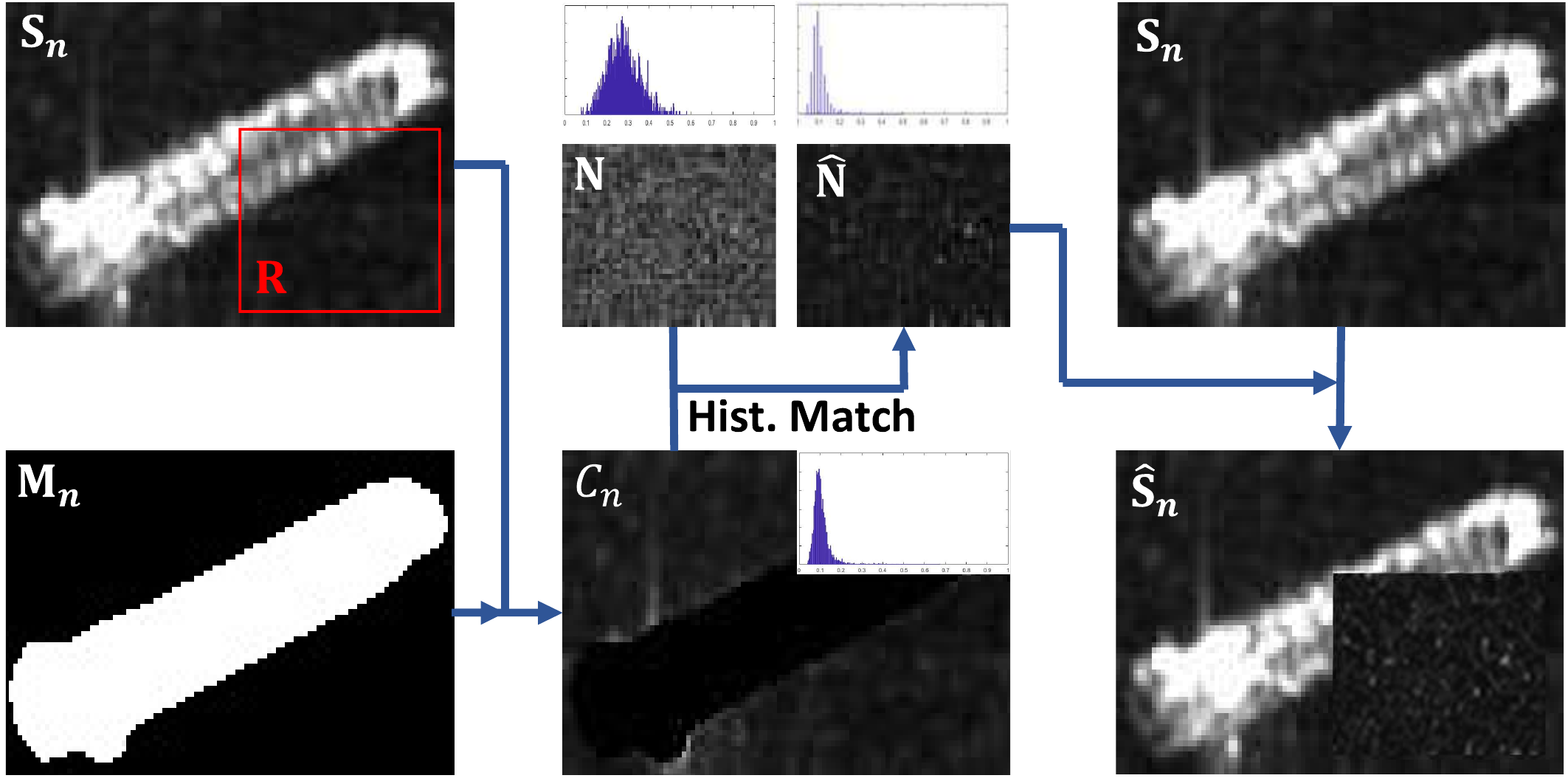}}\vspace{-10pt}
\caption{Pipeline of our proposed context-preserving instance-level augmentation method.}
\vspace{-15pt}
\label{fig:augmentation}
\end{figure}
    \subsection{Context-preserving Instance-level Data Augmentation}\label{sec:proposed_aug}
\begin{figure}[]
\centering
\renewcommand{\thesubfigure}{}
\subfigure[(a) Original]{\includegraphics[width=0.49\columnwidth]{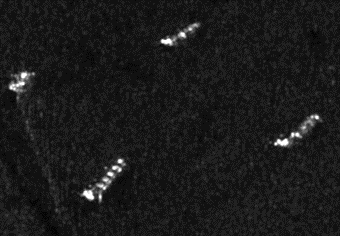}}
\subfigure[(b) Random erasure \cite{zhong2020random}]{\includegraphics[width=0.49\columnwidth]{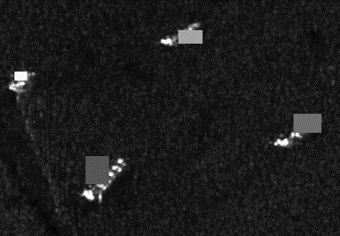}}
\\\vspace{-5pt}
\subfigure[(c) Direct insertion \cite{he2019adaptive, he2020fusion}]{\includegraphics[width=0.49\columnwidth]{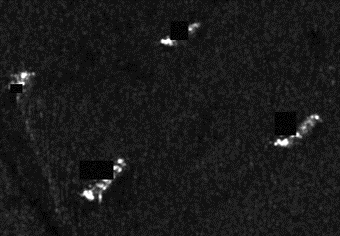}}
\subfigure[(d) Proposed Method]{\includegraphics[width=0.49\columnwidth]{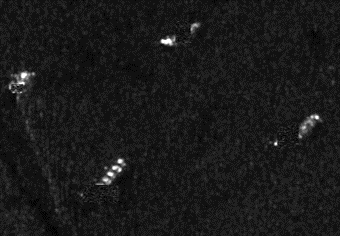}}
\\\vspace{-5pt}
\caption{Example of different random patch replacement methods. (a) original image, (b) random erasure \cite{zhong2020random}, (c) direct insertion with noise patch \cite{he2019adaptive, he2020fusion}, and (d) our proposed method.}
\vspace{-20pt}
\label{fig: badAug}
\end{figure}

        In this section, we present details of the proposed context-preserving instance-level data augmentation method.
        As illustrated in Fig. \ref{fig:shadow}, SAR images often include targets with information loss caused by radar shadows, due to its geometries during acquisition process.
        It causes shape variations of targets and often poses challenges in detecting the targets.
        To tackle this issue, we propose a data augmentation technique that simulates the information loss on each target instance, illustrated in Fig. \ref{fig:augmentation}.
        A straightforward way to simulate information loss is to directly replace a random area inside the bounding box with a random value \cite{zhong2020random}, or a noise patch sampled from background images \cite{he2019adaptive, he2020fusion}.
        However, such methods can result in unnaturally inserted patch and corrupt the contextual information around the targets, as shown in Fig. \ref{fig: badAug}.

        To alleviate this problem, we propose a novel process that utilizes the instance segmentation mask to preserve the contexts.
        We consider a sub-image $\mathbf{S}_n$ inside corresponding ground-truth bounding box $\mathbf{b}_n$ of the $n^{th}$ ship instance.
        Similar to previous approaches~\cite{zhong2020random}, we first define random rectangle $\mathbf{R}$ of size $[w_o, h_o]$ with area ratio parameter $r_S$ and aspect ratio parameter $r_A$ as:
        \begin{equation}
            w_o h_o= r_S(wh),
        \end{equation}
        \begin{equation}
            w_o = h_o/r_A,
        \end{equation}
        such that its area is $r_S$ times of that of $\mathbf{b}_n$, and ratio of its width and height is $r_A$.
        Since radar shadow is likely to be cast towards outer boundary of an target \cite{bolter2000shape}, we add another constraint that at least one edge of the rectangle should be overlapped with an edge of the bounding box.
        Then, we define a set of pixel values $C_n$ as:
        \begin{equation}
            C_n = \{\mathbf{S}_n(\mathbf{x}), ~\text{for}~\mathbf{x}~ \text{such that}~\mathbf{M}_n(\mathbf{x})=0\},
        \end{equation}
        where $\mathbf{x} = [x,y]$ is pixel index.
        It consists of pixel values inside the bounding box but do not belong to the target, and contains the contextual information of adjacent background of each ship instance.
        We randomly sample a noise patch $\mathbf{N}\in\mathbb{R}^{w_o \times h_o}$ from background images and perform histogram matching to that of $C_n$ to obtain $\hat{\mathbf{N}}$.
        By performing histogram matching, the overall distribution of pixel intensities of the noise patch is matched to that of the contextual background.
        Finally, $\hat{\mathbf{N}}$ is inserted into $\mathbf{R}$, generating the augmented sub-image $\hat{\mathbf{S}}_n$.
        Consequently, we effectively simulate information loss to perform instance-level augmentation while preserving the contextual information. 

    \subsection{Shape-adaptive Deep Detector with DCN} 
    \label{sec:proposed_network}
        We consider a standard two-stage deep detector for ship detection.
        It designs region proposal networks (RPN) by sharing convolutional backbone features with the down-stream detection network, improving the quality of region proposal and overall object detection accuracy.
        Although they improve the detection accuracy by learning the region proposal network, the fixed geometric structure of conventional CNNs cannot handle various geometric transformations that widely occur in SAR targets.
        To improve the robustness against target shape variation, we use network architecture that enables shape-adaptive extraction of deep features.
        Specifically, we adopt Deformable Convolutional Network (DCN) \cite{dai2017deformable} within our network.
        
\begin{figure}[t]
\centering
{\includegraphics[width=0.8\columnwidth]{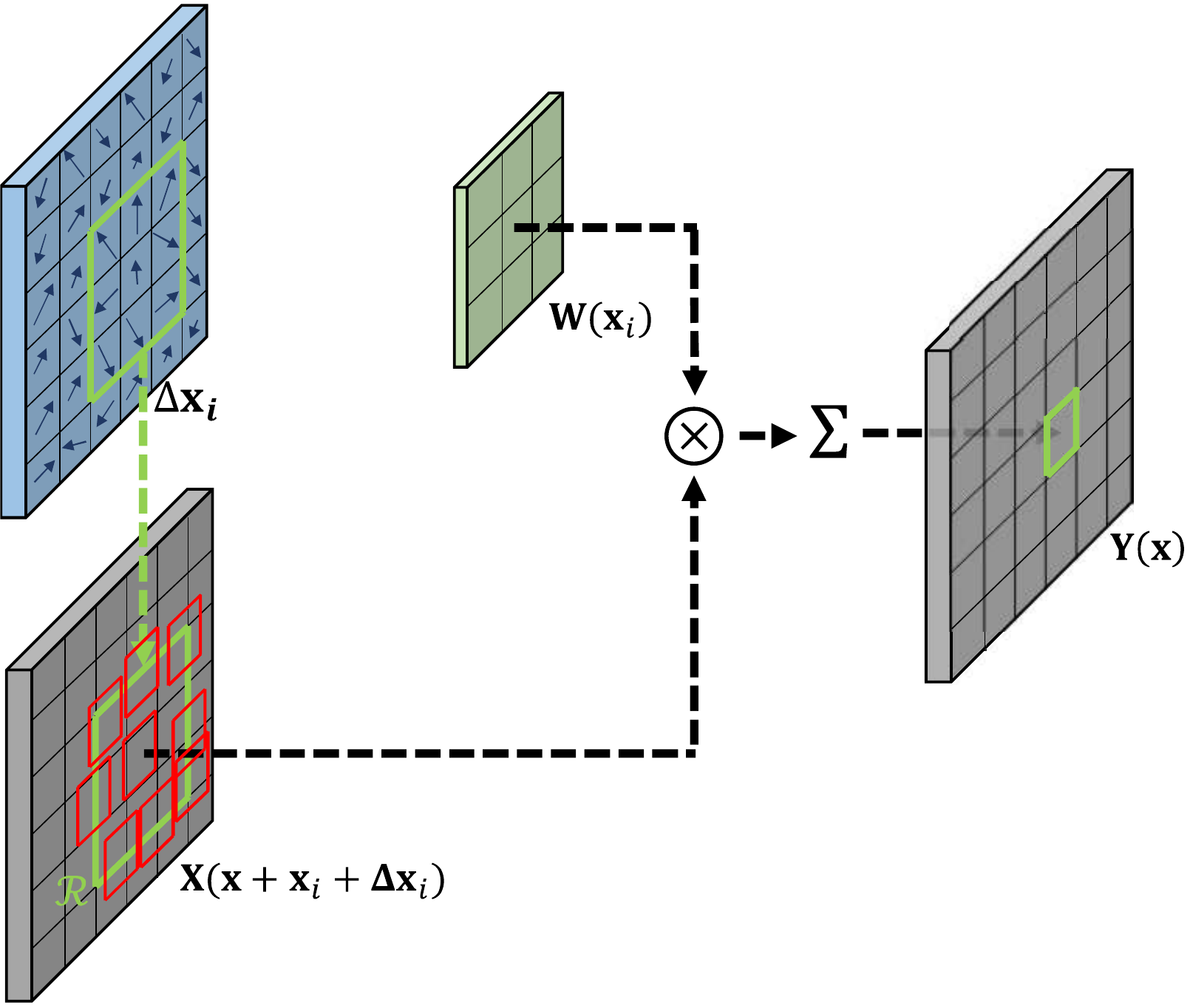}}\vspace{-10pt}
\caption{Illustration of operation of a 3$\times$3 deformable convolution layer.}
\vspace{-15pt}
\label{fig:DCN}
\end{figure}

        DCN consists of two modules, \emph{i.e.}, deformable convolution and deformable RoI pooling.
        As illustrated in Fig. \ref{fig:DCN}, deformable convolution learns 2D offset to the grid of standard rectangular convolution sampling points and extracts convolutional features from geometrically varying area.
        The operation of the deformable convolution is formulated as:
        \begin{equation}
            \mathbf{Y}(\mathbf{x}) = \sum_{\mathbf{x}_i \in \mathcal{R}} \mathbf{W}(\mathbf{x}_i)\mathbf{X}(\mathbf{x}+\mathbf{x}_i+\Delta\mathbf{x}_i),
        \end{equation}
        where $\mathbf{X}$ is input CNN feature, $\mathbf{W}$ is convolutional filter, $\mathbf{x}_i$ is the grid position within the filter's receptive field $\mathcal{R}$, $\Delta\mathbf{x}$ is the sampling offset, and $\mathbf{Y}$ is the output feature.
        Similarly, deformable RoI pooling adds an offset to the RoI extractor as:
        \begin{equation}
            \mathbf{Y}(\mathbf{x}) = \sum_{\mathbf{x}_i \in \text{bin}(\mathbf{x})} \mathbf{X}(\mathbf{x}_i+\Delta\mathbf{x}_i) / N_\text{bin},
        \end{equation}
        where $\text{bin}(\mathbf{x})$ indicates region of the input feature $\mathbf{X}$, allocated to pixel $\mathbf{x}$ of the pooled feature $\mathbf{Y}$, and $N_{bin}$ is number of pixels in the $\text{bin}(\bf{x})$.
        It enables adaptive part localization for objects with different shape and enhances flexibility in reflecting shape variation in each object.

\section{Experiments}
    \subsection{Experimental Settings}\label{sec:exp_setting}
        \noindent\textbf{Dataset.}
            We conduct experiments on the HRSID dataset \cite{wei2020hrsid}.
            It consists of 5,604 high-resolution SAR images with size of $800\times800$, divided into 3,642 training and 1,962 testing images. 
            Training and testing sets contain 11,047 and 5,922 annotations for ships, respectively. 
            Each annotation contains ground-truth for bounding box and instance segmentation mask.
            It additionally provides 400 images without ship, from which we manually choose 50 images of flat area and use them to sample the noise patch $\mathbf{N}$ for the proposed  augmentation method. \vspace{-5pt}\\\\
        \textbf{Implementation Details.}
            During training, we randomly choose the data augmentation parameters $r_S\in[0.2, 0.4]$ and $r_A\in[0.5, 2.0]$ for each instance.
            For the testing, the original images are used.
            We use Faster R-CNN \cite{ren2015faster} as baseline detector, with ResNet-50 \cite{he2016deep} Feature Pyramid Network (FPN) \cite{lin2017feature} backbone, initialized with the ImageNet \cite{krizhevsky2012imagenet} pretrained weights.
            For the deep detector with DCN, we apply deformable convolution to the $3\times 3$ convolutional layers in the 3rd through 5th convolutional blocks in ResNet-50, and use deformable RoI pooling.
            During training, the network is optimized using stochastic gradient descent with fixed learning rate 0.01, momentum 0.9, and weight decay 0.0001.
            In addition to the proposed augmentation method, we perform random flip augmentation, and train the network for 60 epochs with batch size 8.
            We use MMDetection \cite{mmdetection} toolbox, built upon Pytorch \cite{pytorch} library.
            All experiments are conducted on a PC with 3.60GHz CPU, 24GB RAM, and a NVIDIA TITAN RTX GPU.\vspace{-5pt}\\\\
        \textbf{Evaluation Metric.}
            To quantitatively evaluate the detection performance, we use a general metric used in object detection, Average Precision (AP), calculated over different IoU threshold or object size.
            More specifically, `$\text{AP}$' indicates average precision over IoU thresholds [0.50:0.05:0.95], `$\text{AP}_{50}$' and `$\text{AP}_{75}$' means precision at IoU = 0.50 and IoU= 0.75, respectively.
            We also separately measure AP scores for small ($\text{AP}_\text{S}$), medium ($\text{AP}_\text{M}$) and large ($\text{AP}_\text{L}$) objects, whose sizes are determined by bounding box area in term of pixels ($\text{S} < 32^2 < \text{M} < 96^2 < \text{L}$).

    \begin{figure*}[t]
	\centering
	\renewcommand{\thesubfigure}{}
    \subfigure[]{\includegraphics[width=0.245\textwidth]{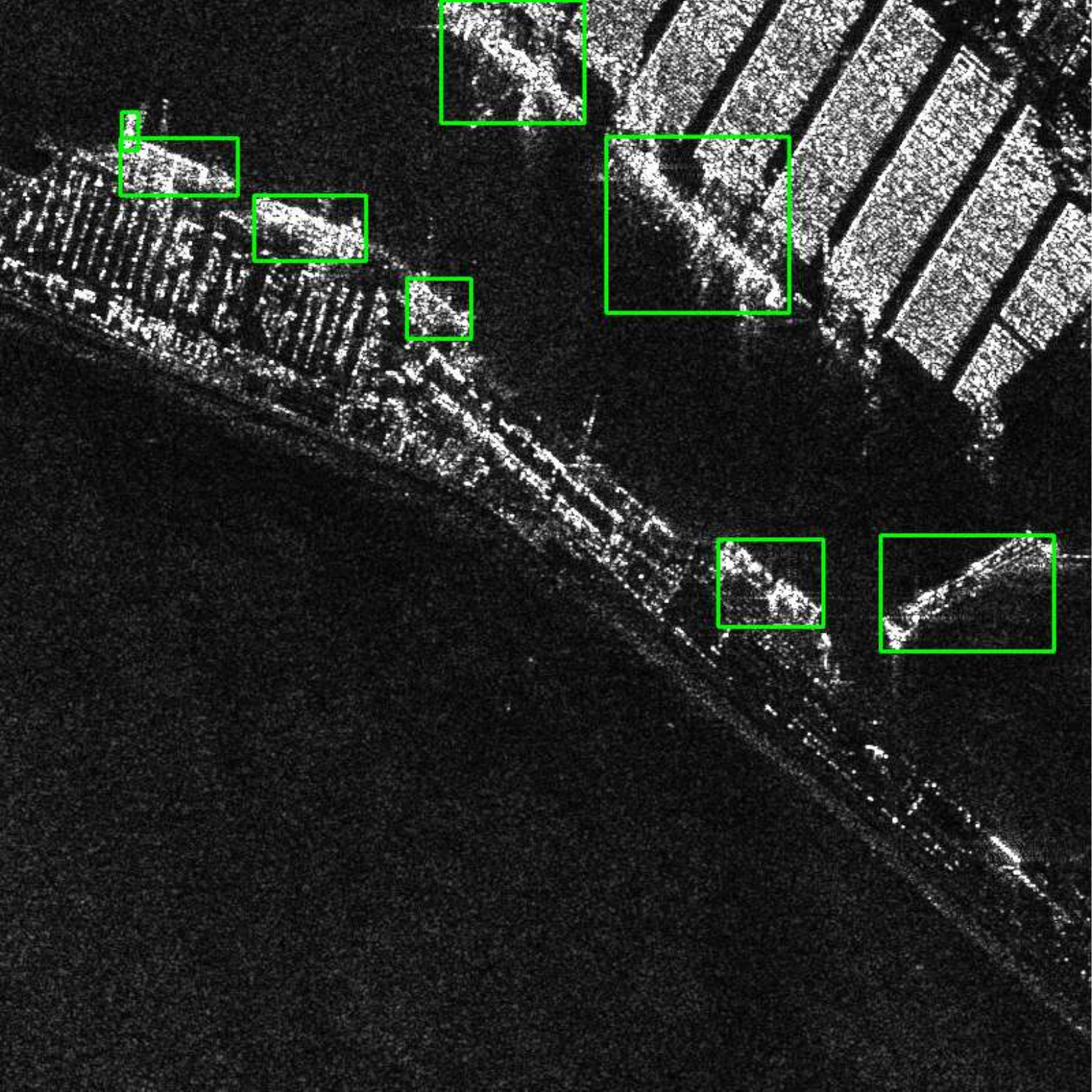}}
    \subfigure[]{\includegraphics[width=0.245\textwidth]{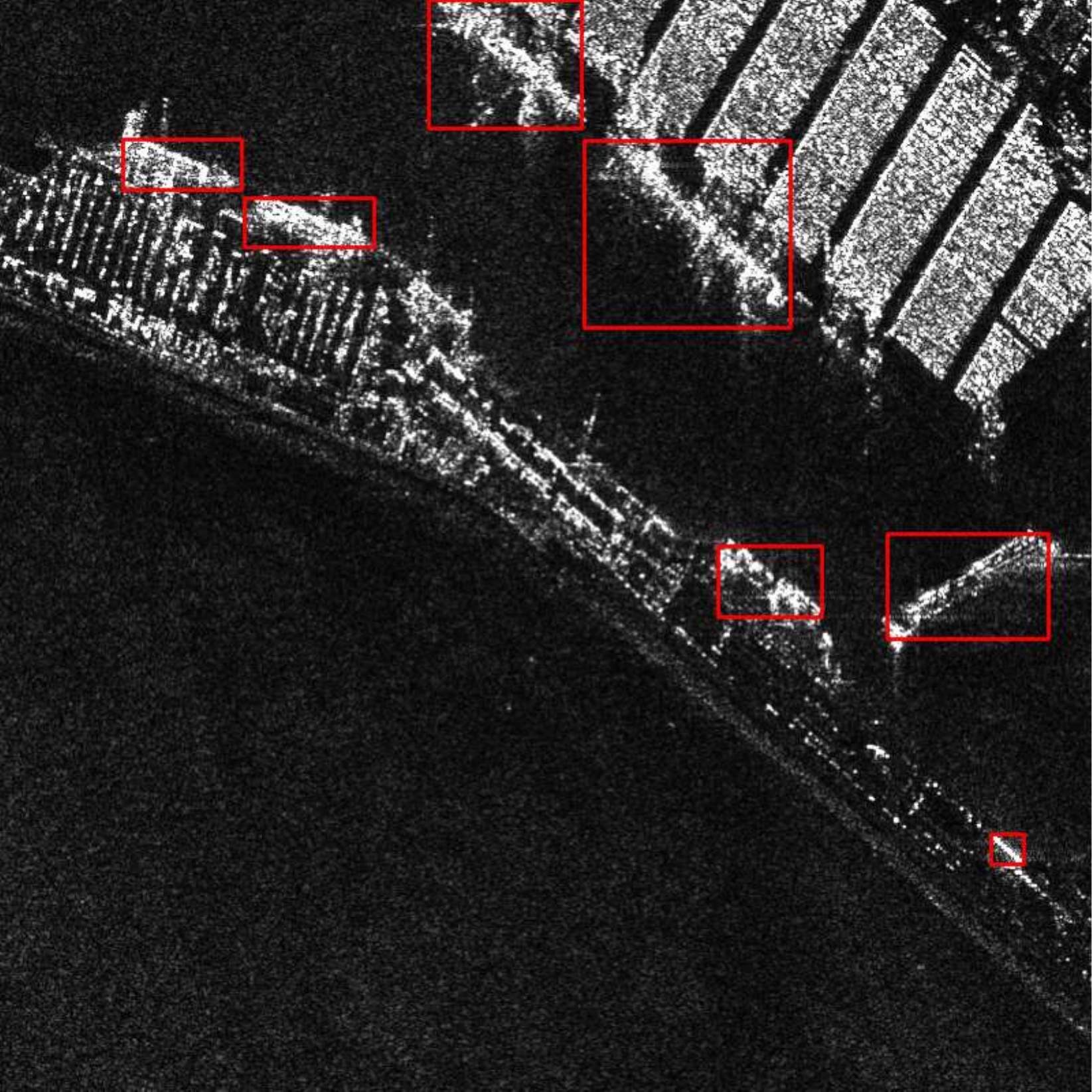}}
    \subfigure[]{\includegraphics[width=0.245\textwidth]{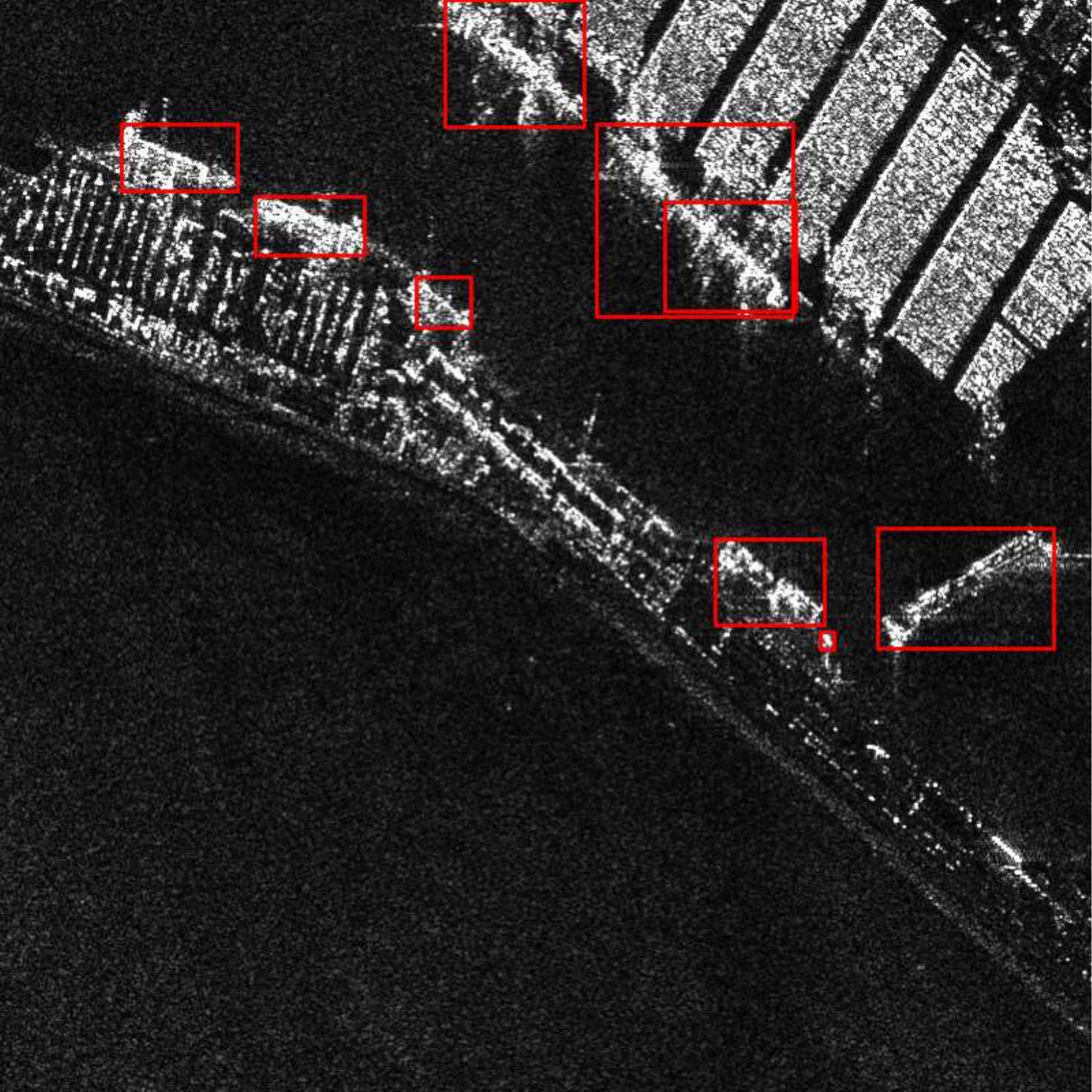}}
    \subfigure[]{\includegraphics[width=0.245\textwidth]{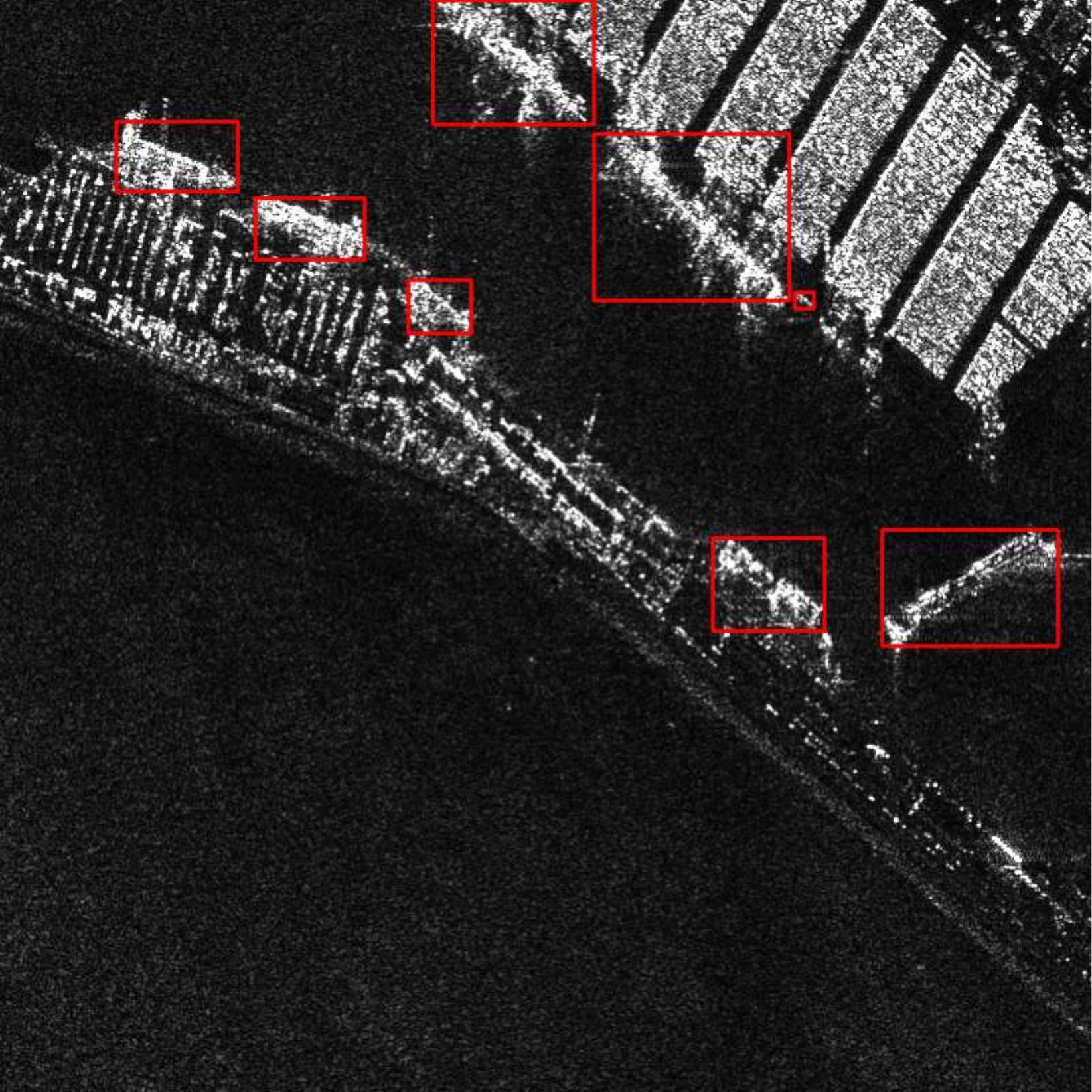}}
    \\\vspace{-15pt}
    \subfigure[(a) Ground-truth]{\includegraphics[width=0.245\textwidth]{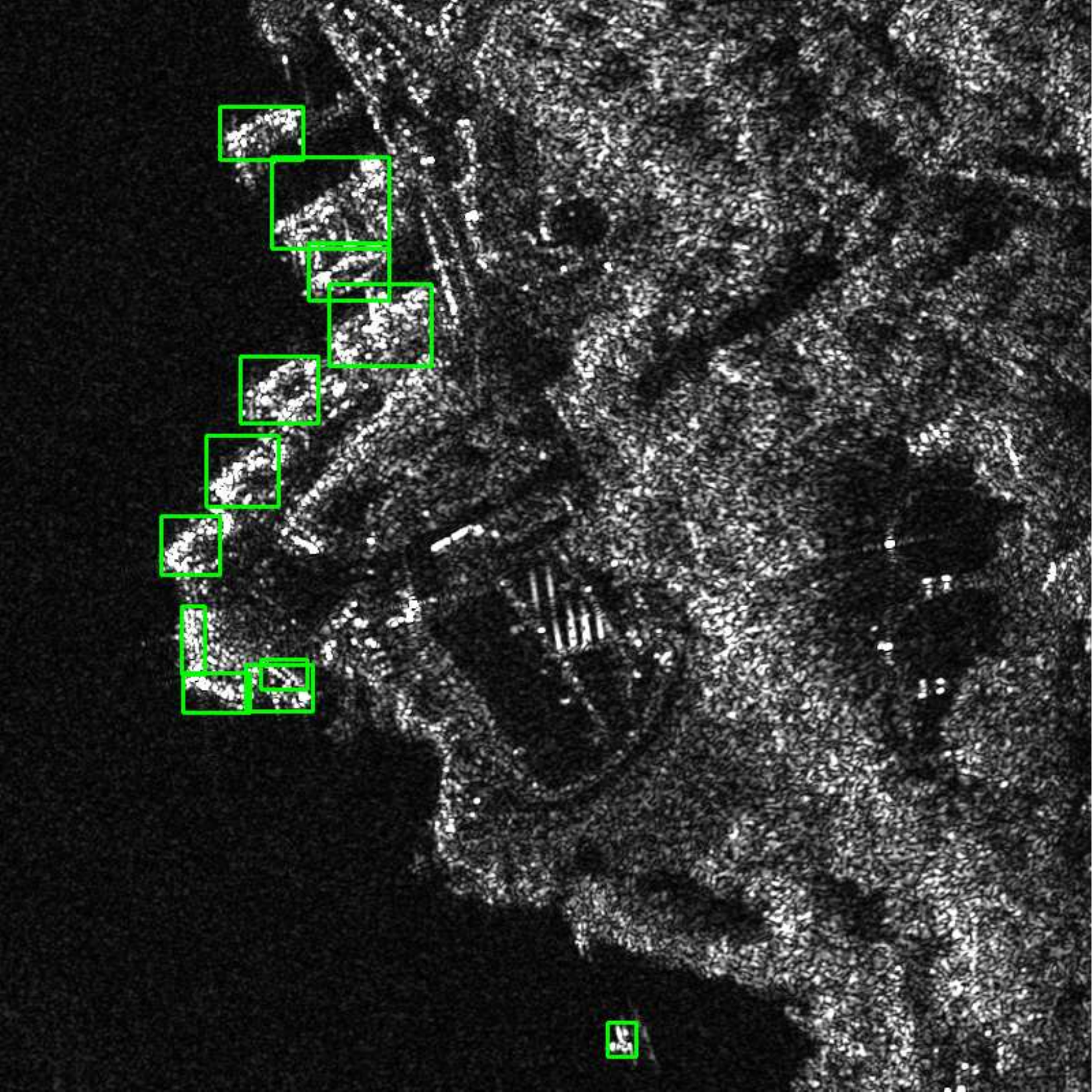}}
    \subfigure[(b) YOLO v3 \cite{farhadi2018yolov3}]{\includegraphics[width=0.245\textwidth]{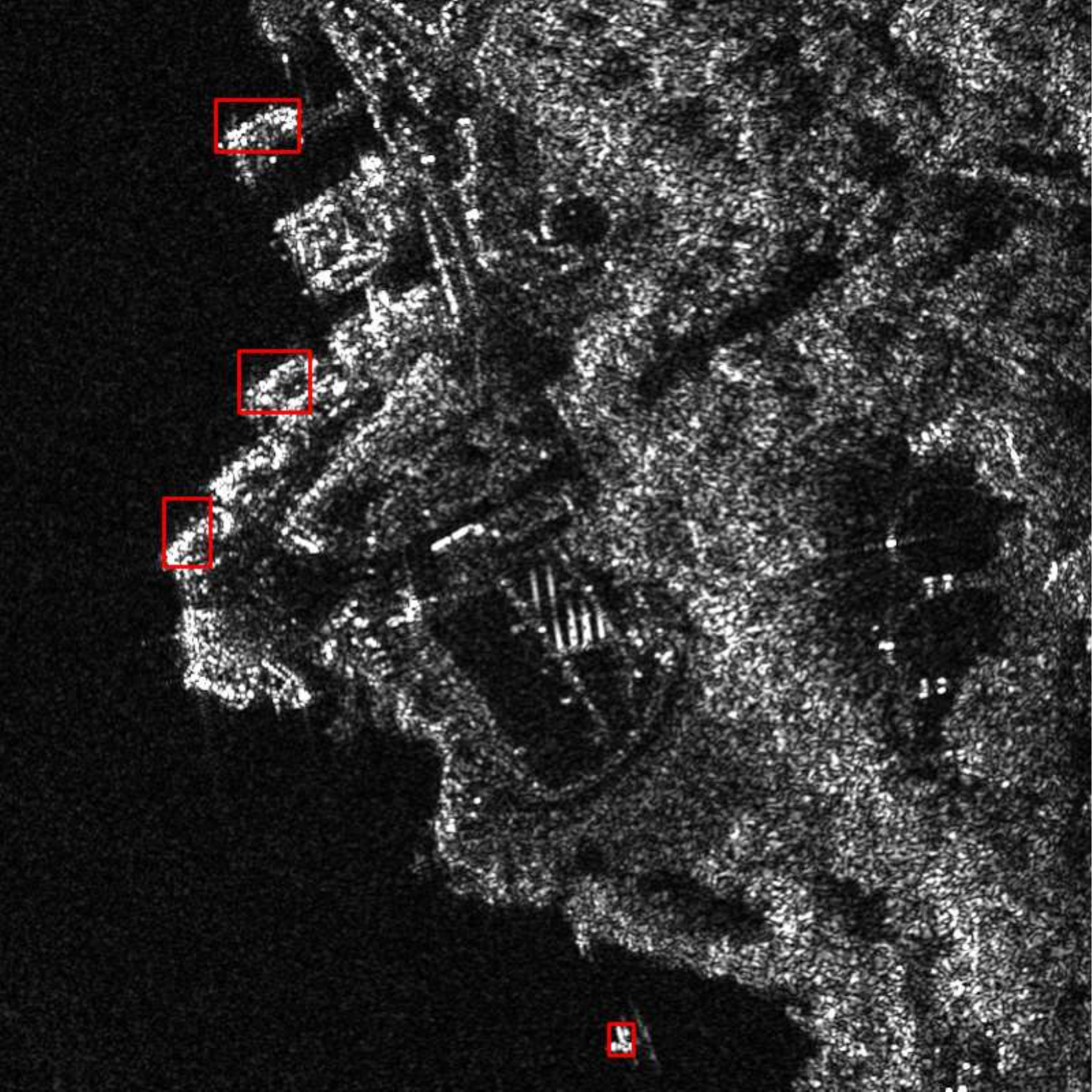}}
    \subfigure[(c) Faster R-CNN \cite{ren2015faster}]{\includegraphics[width=0.245\textwidth]{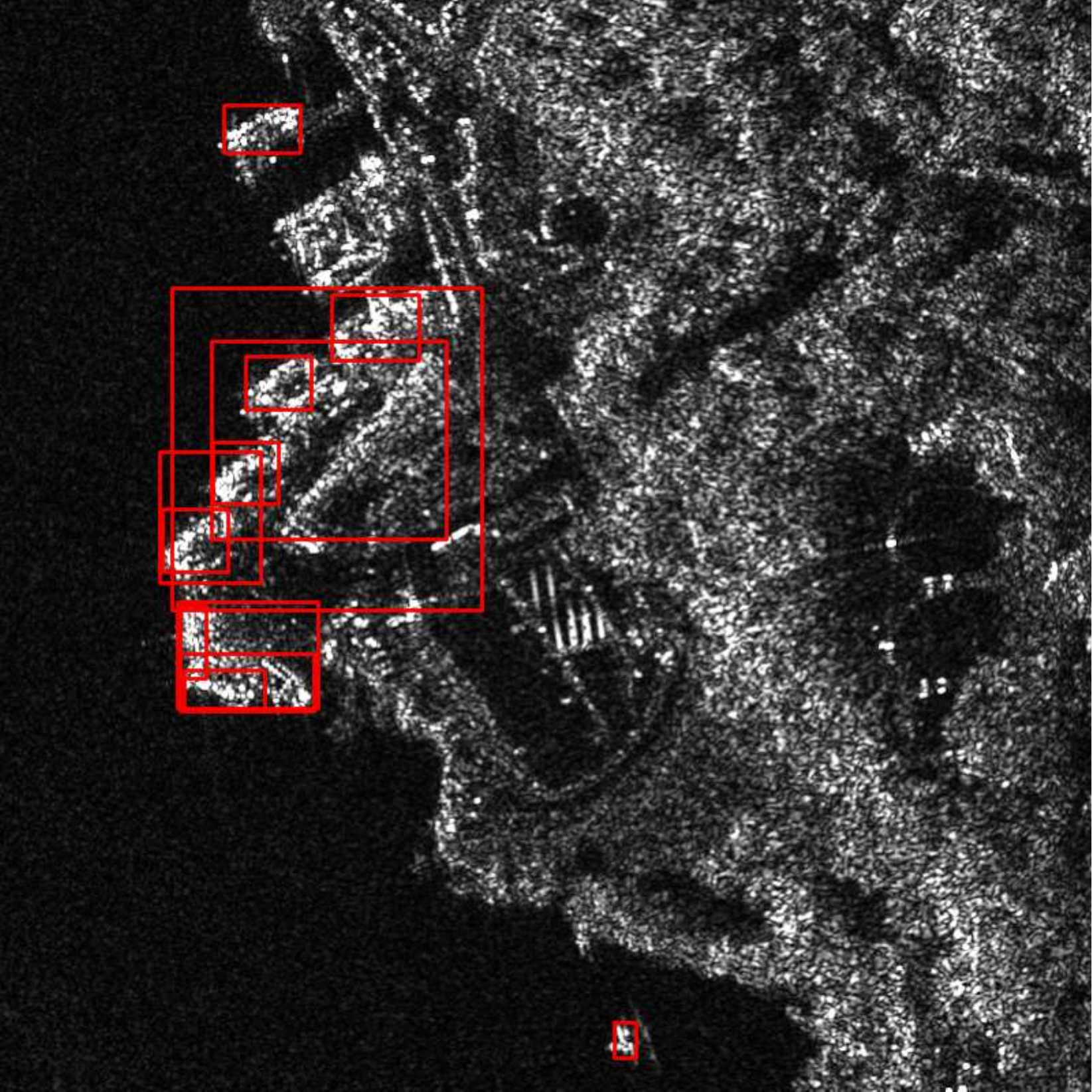}}
    \subfigure[(d) Proposed Method]{\includegraphics[width=0.245\textwidth]{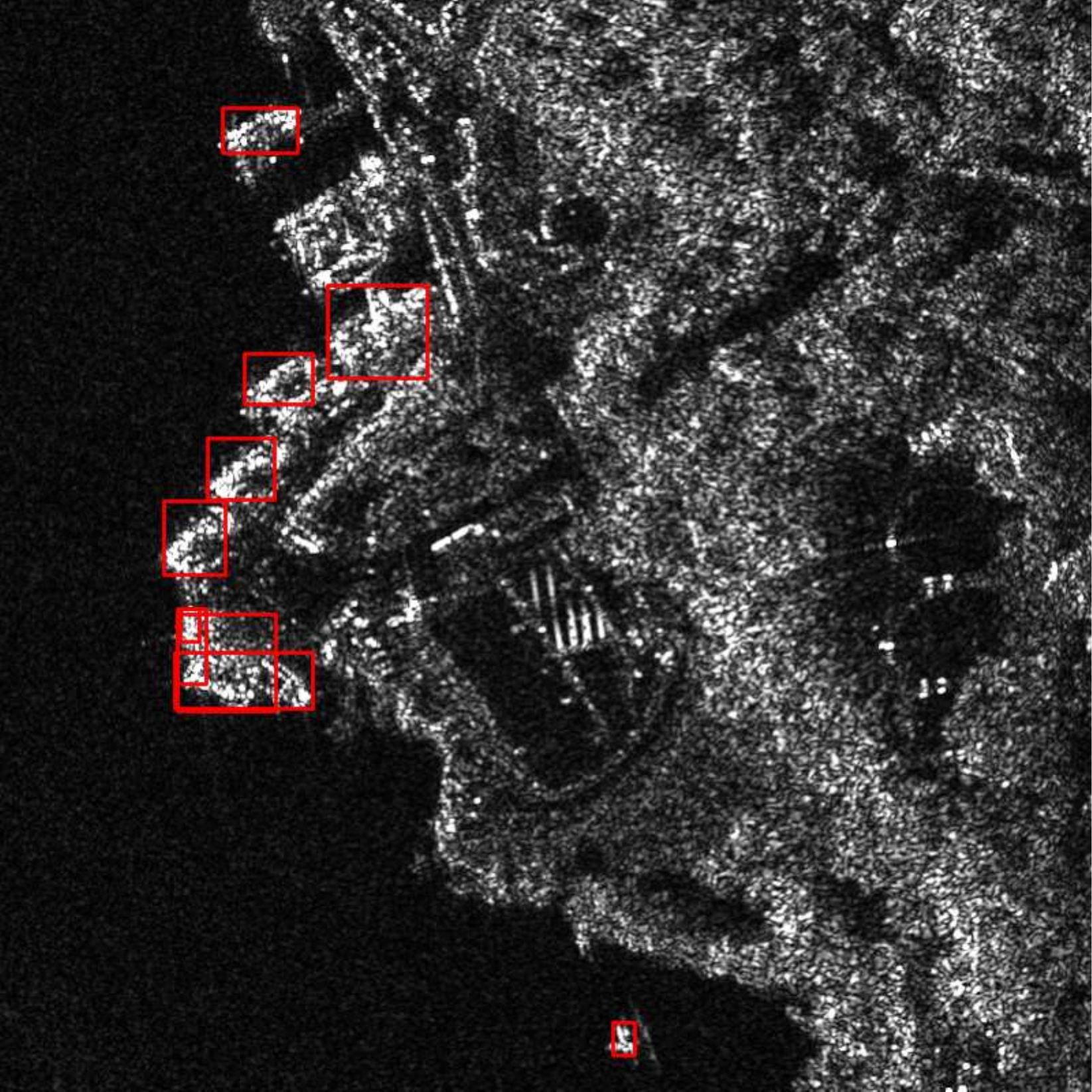}}
    \\\vspace{-5pt}
	\caption{Ship detection results of different methods on the dataset \cite{wei2020hrsid}. (a) ground-truth bounding box, detection results of (b) YOLO v3 \cite{farhadi2018yolov3}, (c) Faster R-CNN \cite{ren2015faster} and (d) proposed method.}
	\vspace{-15pt}
	\label{fig:comparison}
\end{figure*}  
    \subsection{Results}
    \noindent\textbf{Comparisons with other deep detectors.} 
    In this experiment, we compare performances with other one-stage detectors: SSD \cite{liu2016ssd}, YOLO v3 \cite{farhadi2018yolov3}, RetinaNet \cite{lin2017focal}, and two-stage detectors: Faster R-CNN \cite{ren2015faster}, and Mask R-CNN \cite{he2017mask}.
    For these networks, we do not use DCN or data augmentation.
    Other components such as backbone and training hyper-parameters are set as same as ours presented in Sec. \ref{sec:exp_setting}.

\begin{table}[] 
	\caption{Quantitative results of the proposed method and previous deep detectors on the HRSID dataset \cite{wei2020hrsid}.} \vspace{-20pt}
	\begin{center} \label{tab:comparison_net}
	\resizebox{1.00\columnwidth}{!}{
		\begin{tabular}{ccccccc}
			\toprule
			Methods & $\text{AP}$ &$\text{AP}_{50}$ & $\text{AP}_{75}$& $\text{AP}_\text{S}$& $\text{AP}_\text{M}$ & $\text{AP}_\text{L}$\\
			\midrule \midrule
			 SSD \cite{liu2016ssd}& 0.568& 0.849 &0.366 &0.580 &0.578 & 0.170 \\
			 YOLOv3 \cite{farhadi2018yolov3}&  0.517 &0.857 &0.547& 0.520 &0.557 &\bf{0.265} \\
			 RetinaNet \cite{lin2017focal}& 0.538 & 0.797 & 0.594 & 0.546 &  0.606& 0.226 \\
			 \midrule
			 Faster R-CNN \cite{ren2015faster} & 0.619 & 0.897 & 0.711 & 0.636 & 0.602 &0.105  \\
			 Mask R-CNN \cite{he2017mask}& 0.627 & 0.888 & 0.722& 0.649 & 0.587 & 0.201  \\
			 \midrule
			 Proposed Method & \bf{0.664} & \bf{0.914} & \bf{0.766}&\bf{0.680} &\bf{0.645} &0.199 \\
			\bottomrule
		\end{tabular}
		}\vspace{-15pt}
	\end{center}
\end{table}
	
    We present the quantitative results in Table \ref{tab:comparison_net}.
    The one-stage detectors, SSD (0.568 AP), YOLO v3 (0.517 AP), and RetinaNet (0.538 AP), show lower detection performance in general.
    The two-stage detectors, Faster R-CNN (0.619 AP), and Mask R-CNN (0.627 AP), show better performance compared to the one-stage detectors.
    Especially, they show noticeable improvements in $\text{AP}_{75}$, indicating better localization performance.
    Trained with more diverse data and having enhanced capacity in modeling geometric variation with DCN, our proposed method achieves highest AP of 0.664 compared to other methods, and also shows superior results for other metrics, except $\text{AP}_\text{L}$.
    Also, with the qualitative results in Fig. 7, we observe that our proposed method generates detection results with less missing targets and false positives compared to the other networks.
    \vspace{-5pt}\\\\
    \noindent\textbf{Comparisons with other augmentation methods.} 
    We compare the effect of different data augmentation methods on the detection performance.
    To this end, we train Faster R-CNN \cite{ren2015faster} with DCN, using datasets i) without augmentation, ii) Random Erasure (RE) \cite{zhong2020random}, iii) Direct Background Insertion (DBI) \cite{he2019adaptive,he2020fusion}, and iv) our proposed method.
    For all methods, augmentation parameters $r_S$ and $r_A$ are set as described in Sec. \ref{sec:exp_setting}.
\begin{table}[]
	\caption{Detection results of Faster R-CNN \cite{ren2015faster} with different training data configurations.} \vspace{-20pt}
	\begin{center} \label{tab:comparison_aug}
	\resizebox{1.00\columnwidth}{!}{
		\begin{tabular}{ccccccc}
			\toprule
			DA & $\text{AP}$ &$\text{AP}_{50}$ & $\text{AP}_{75}$& $\text{AP}_\text{S}$& $\text{AP}_\text{M}$ & $\text{AP}_\text{L}$\\
			\midrule \midrule
			None & 0.651 &0.892 &0.756 &0.671 &0.622 &0.147  \\
			RE \cite{zhong2020random} &0.655 &0.907 &0.764 &0.671& 0.633 &0.145   \\
			DBI \cite{he2019adaptive, he2020fusion} &0.657& 0.905& 0.757& 0.675& 0.639& \bf{0.205} \\
			Proposed &\bf{0.664} & \bf{0.914} & \bf{0.766}&\bf{0.680} &\bf{0.645} &0.199  \\
			 \bottomrule
		\end{tabular}}
	\end{center}\vspace{-20pt}
\end{table}
	
    The quantitative results are presented in Table \ref{tab:comparison_aug}. 
    Compared to the network trained with original data without augmentation, all the augmentation methods result in improved AP.
    We observe network trained with RE \cite{zhong2020random} augmentation shows slightly degraded performance in detecting large objects (0.145 $\text{AP}_\text{L}$), compared to the baseline (0.147 $\text{AP}_\text{L}$).
    DBI \cite{he2019adaptive,he2020fusion} achieves best score for $\text{AP}_\text{L}$, which is comparative to our proposed method.
    The all the other metrics, our proposed method yields in highest scores.
    \vspace{-5pt}\\\\
    \noindent\textbf{Ablation Study.}
    Here, we perform ablation study to observe effects of each proposed component.
    Using Faster R-CNN \cite{ren2015faster} as baseline, we train the network using different data augmentation and DCN configurations. 
    We train the baseline network with dataset without augmentation, and the proposed context-preserving instance-level augmentation.
    Then, we adopt DCN and repeat the experiment to observe its effect on detection performance.
    
    Quantitative results are presented in Table \ref{tab:Ablation}.
    It is observed that applying the proposed data augmentation method improves the overall detection performance (0.630 AP) of the baseline (0.619 AP).
    Applying DCN results in further increased AP for all data configurations, achieving 0.651 AP without augmentation, and 0.664 AP with the proposed augmentation method.
    Consequently, we find that both the proposed context-preserving instance-level data augmentation and deep detector with DCN contribute to improving the detection performance.
    
\begin{table}[]
	\caption{Quantitative results of ablation study.} \vspace{-20pt}
	\begin{center} \label{tab:Ablation}
	\resizebox{1.00\columnwidth}{!}{
		\begin{tabular}{cccccccc}
			\toprule
			DA & DCN & $\text{AP}$ &$\text{AP}_{50}$ & $\text{AP}_{75}$& $\text{AP}_\text{S}$& $\text{AP}_\text{M}$ & $\text{AP}_\text{L}$\\
			\midrule \midrule
			 &  & 0.619 & 0.897 & 0.711 & 0.636 & 0.602 &0.105  \\
			$\surd$ &  & 0.630 & 0.905 & 0.723&0.646 &0.633 & \bf{0.202}  \\
			 & $\surd$ & 0.651 &0.892 &0.756 &0.671 &0.622 &0.147  \\
			$\surd$ & $\surd$ &\bf{0.664} & \bf{0.914} & \bf{0.766}&\bf{0.680} &\bf{0.645} &0.199  \\
			 \bottomrule
		\end{tabular}}
	\end{center}\vspace{-20pt}
\end{table}
\section{Conclusion}
In this paper, we propose context-preserving instance-level data augmentation method to build a deep SAR ship detection system that is robust to target shape variations.
Taking advantage of ground-truth information for bounding box and instance segmentation mask, we effectively simulate the information loss within targets to emulate occlusion or radar shadow.
We also adopt deformable convolutional network to enhance capacity of the network in modeling geometric variations. 
We perform extensive experiments on the HRSID dataset and observe improved ship detection performance of the proposed method.
\section{Acknowledgements}
This work was supported by a grand-in-aid of Hanwha Systems.

\bibliographystyle{IEEEtran}
\bibliography{refs}
\end{document}